\documentclass[runningheads]{llncs}
\usepackage{graphicx}

\usepackage{tikz}
\usepackage{comment}
\usepackage{amsmath,amssymb} 
\usepackage{color}
\usepackage{wrapfig}

\usepackage[accsupp]{axessibility}  

\usepackage{epsfig}
\usepackage{graphicx}
\usepackage{multirow}
\usepackage{subcaption}
\usepackage{tabularx}
\usepackage{floatrow}
\newfloatcommand{capbtabbox}{table}[][\FBwidth]

\usepackage{blindtext}

\usepackage{enumitem}

\begin{document}
\pagestyle{headings}
\mainmatter
\def\ECCVSubNumber{3685}  

\title{VecGAN: Image-to-Image Translation with Interpretable Latent Directions} 


\titlerunning{VecGAN}
%
\author{Yusuf Dalva \and
Said Fahri Altındiş \and
Aysegul Dundar}
\authorrunning{Y. Dalva et al.}
%
\institute{Bilkent University \\
\email{\{yusuf.dalva, fahri.altindis\}@bilkent.edu.tr}\\
\email{adundar@cs.bilkent.edu.tr}}
\maketitle

\begin{abstract}
We propose VecGAN, an image-to-image translation framework for facial attribute editing with interpretable latent directions.
Facial attribute editing task faces the challenges of precise attribute editing with controllable strength and preservation of the other attributes of an image.
For this goal, we design the attribute editing by latent space factorization and for each attribute, we learn a linear direction that is orthogonal to the others.
The other component is the controllable strength of the change, a scalar value.
In our framework, this scalar can be either sampled  or encoded from a reference image by projection.
Our work is inspired by the latent space factorization works of fixed pretrained GANs.
However, while those models cannot be trained end-to-end and struggle to edit encoded images precisely, VecGAN is end-to-end trained for image translation task and successful at editing an attribute while preserving the others. 
Our extensive experiments show that VecGAN achieves significant improvements over state-of-the-arts for both local and global edits. 
\end{abstract}

\section{Introduction}
There has been a significant progress in image-to-image translation methods  \cite{isola2017image,park2019semantic,yi2019apdrawinggan,dundar2020panoptic,liu2016coupled,yi2017dualgan,liu2017unsupervised,mardani2020neural} especially for facial attribute editing \cite{starganv2,shen2017learning,xiao2018elegant,zhang2018generative,li2021image} powered with generative adversarial networks (GANs).
A main challenge of facial attribute editing methods is to be able to change only one attribute of an image without affecting others such as global lighting parameters of the images,  identity of the persons, background, or their other attributes.
The other challenge is the interpretability of the style codes so that one can control the attribute intensity of the edit, e.g. increase the intensity of smile or aging.

To achieve the targeted attribute editing while preserving the others, many works set a separate style encoder and an image editing network where modified styles are injected into it \cite{starganv2,li2021image}.
During image-to-image translation, a style encoded from another image or a newly sampled style latent code can be used to output diverse images.
To disentangle attributes, works focus on style encoding and progress from a shared style code, SDIT \cite{wang2019sdit}, to mixed style codes, StarGANv2 \cite{starganv2}, to hierarchical disentangled styles, HiSD \cite{li2021image}.
Among these works, HiSD independently learn styles of each attribute, bangs, hair color, glasses and introduces a local translator which uses attention masks to avoid global manipulations.
HiSD showcases successes on those three local attribute editing task and is not tested for global attribute editing, e.g. age, smile.
Furthermore, one limitation of these works is the uninterpretablity of style codes as one cannot control the intensity of attribute (e.g. blondness) in a straight-forward manner.

To overcome the challenges of facial attribute editing task, we propose a novel framework, VecGAN, and image-to-image translation framework with interpretable latent directions. 
Our framework does not require a separate style encoder  as in the previous works since we achieve the translation in the encoded latent space directly.
The attribute editing directions are learned in the latent space and regularized to be orthogonal to each other for style disentanglement.
The other component of our framework is the controllable strength of the change, a scalar value.
This scalar can be either sampled from a distribution or encoded from a reference image by projection in the latent space.
Our framework not only achieves significant improvements over state-of-the-arts for both local and global edits but also provides a knob to control the editing attribute intensity via its design.

\begin{figure}[t]
    \centering
    \includegraphics[width=0.9\textwidth]{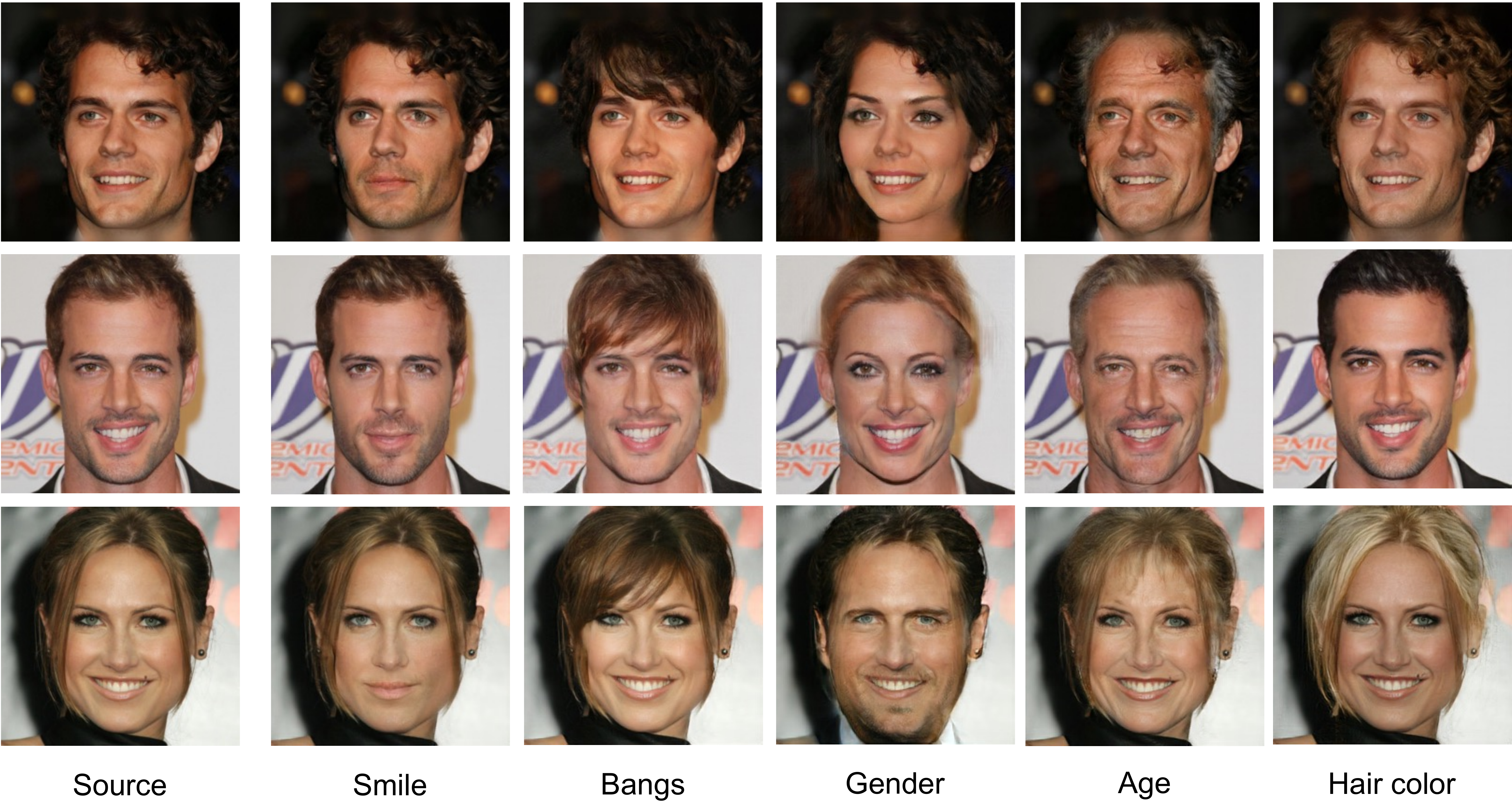}
    \caption{Attribute editing results of VecGAN.
    The first column shows the source images, and other columns show the results of editing a specific attribute. Each edited image has an attribute value opposite to that of the source one.
    For hair color, sources are translated to brown, black, and blonde hair, respectively.}
    \label{fig:all_samples}
\end{figure}

VecGAN is encouraged by the findings that well-trained generative models organize their latent space as disentangled  representations with meaningful directions in a completely unsupervised way.
Exploring these interpretable directions in latent codes has emerged as an important research endeavor on the fixed pretrained GANs \cite{shen2020interpreting,voynov2020unsupervised,harkonen2020ganspace,shen2021closed,wu2021stylespace}.
These works show that images can be mapped to the GANs latent space and edits can be achieved by manipulations in the latent space.
However, since these models are not trained end-to-end, the results are sub-optimal as will also be shown in our experiments.

To enable VecGAN, different than previous works of image-to-image translation networks, we use a deeper neural network architecture.
Image-to-image translation methods, such as state-of-the-art HiSD \cite{li2021image} uses a network with small receptive fields that decreases the image resolution only by four times in the encoder.
However, we want an organization in a latent space such that we can take meaningful linear directions.
Therefore, images should be encoded to a spatially smaller feature space and a network should have a full understanding of an image.
For that reason, we set a deep encoder and decoder network architecture but then this
network faces the challenges of reconstructing all the details from the input image.
To solve this problem, we use a skip connection  between  the  encoder  and  decoder but only at lower resolution to find the optimal equilibrium of the information flow between with and without dimensionality reduction bottleneck.
In summary, our main contributions are:
\begin{itemize}[leftmargin=*]
    \item We propose VecGAN, a novel image-to-image translation network that is trained end to end with interpretable latent directions. Our framework does not employ a separate style network as in the previous works and translations are achieved with a single deep encoder-decoder architecture. 
    \item VecGAN enables both reference attribute copy and attribute strength manipulation. 
    Reference style encoding is designed in a novel way by using the same encoder from the translation pipeline.
    First, encoder is used to obtain latent codes of a reference image and it is followed by the projection of the codes into learned latent directions for different attributes.
    \item We conduct extensive experiments to show the effectiveness of our framework and achieve significant improvements over state-of-the-art for both local and global edits. Qualitative results of our framework can be seen in Fig. \ref{fig:all_samples}.
\end{itemize}

\section{Related Works}

\textbf{Image to Image Translation.}
Image-to-image translation algorithms aim at preserving a given content while changing targeted attributes.
Examples range from translating semantic maps into RGB images \cite{wang2018high}, to translating summer images into winter images \cite{isola2017image}, to portrait drawing \cite{yi2020unpaired}  and very popularly to editing faces \cite{starganv2,shen2017learning,xiao2018elegant,zhang2018generative,li2021image,wu2019relgan,gao2021high,hou2022guidedstyle,abdal2021styleflow}.   
These algorithms powered with GAN loss \cite{goodfellow2014generative} set an encoder-decoder architecture.
In models that learn a deterministic mapping from one domain to the other, images are processed with encoder and decoder to output translated images  \cite{wang2018high,park2019semantic}.
In multi-modal image-to-image translation methods, style is encoded separately from an another image or sampled from a distribution \cite{huang2018multimodal,starganv2}.
In the generator, style and content are either combined with concatenation \cite{zhu2017multimodal}, or combined with a mask \cite{li2021image} or fed separately through instance normalization blocks \cite{huang2018multimodal,zhu2020sean}.
The generator also uses an encoder-decoder architecture \cite{li2021image,yang2021l2m} that is seperate than the style encoder.
In our work, we are interested in designing the attribute as a learnable linear direction in the latent space and we do not employ a separate style encoder which results in a more intuitive framework.

\textbf{Learning interpretable latent directions.} In another line of research, it is shown that GANs that are trained to synthesize faces can also be used for face attribute manipulations \cite{karras2019style,brock2018large,karras2020analyzing}.
Initially, these networks are not designed or trained to translate images but rather to synthesize high fidelity images. 
However, it is shown that one can embed existing images into the GAN's embedding space \cite{abdal2019image2stylegan} and further one can find latent directions to edit those images \cite{shen2020interpreting,voynov2020unsupervised,harkonen2020ganspace,shen2021closed,wu2021stylespace}.
These directions are explored in supervised \cite{shen2020interpreting} and unsupervised ways \cite{voynov2020unsupervised,harkonen2020ganspace,shen2021closed,wu2021stylespace}.
It is quite remarkable when the generative network is only taught to synthesize realistic images, it organizes the use of latent space such that linear shifts on them change a specific attribute.
Inspired by these findings, we design our image to image translation such that a linear shift in the encoded features is expected to change a single attribute of an image.
Different than previous works, our framework is trained end-to-end for translation task and allows for reference guided attribute manipulation via projection.

\section{Method}

We follow the hierarchical labels defined by \cite{li2021image}.
For a single image, its attribute for tag $i \in \{1,2,...,N\}$ can be defined as $j \in \{1,2,...,M_i\}$, where N is the number of tags and $M_i$ is the number of attributes for tag $i$.
For example $i$ can be tag of hair color, and attribute $j$ can take the value of black, brown, or blond.

Our framework has two main objectives. As the main task, we aim to be able to perform the image-to-image translation task in a feature (tag) specific manner. While performing this translation, as the second objective, we also want to obtain an interpretable feature space which allows us to perform tag-specific feature interpolation.


\subsection{Generator Architecture}
\begin{figure}[t]
    \centering
    \includegraphics[width=1.0\textwidth]{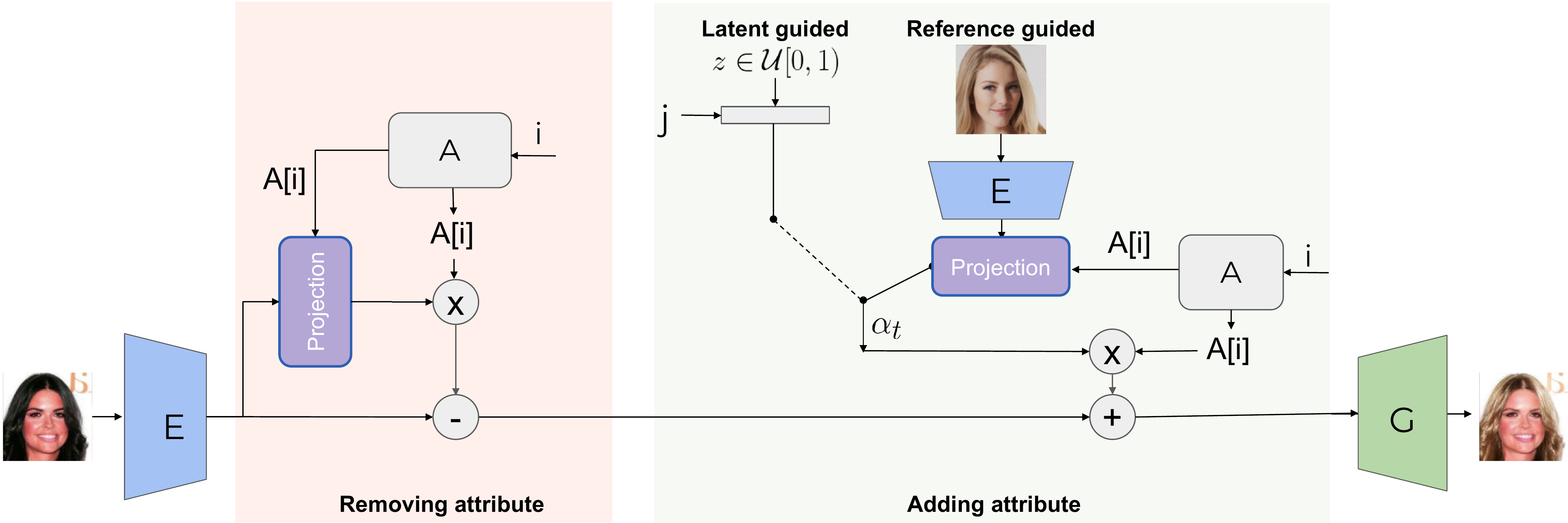}
    \caption{\textbf{VecGAN pipeline.} Our translator is built on the idea of interpretable latent directions. We encode images with an Encoder to a latent representation from which we change a selected tag ($i$), e.g. hair color with a learnable direction $A_i$ and a scale $\alpha$. To calculate the scale, we subtract the target style scale from the source style.
    This operation corresponds to removing an attribute and adding an attribute.
    To remove the image's attribute, source style is encoded and projected from the source image.
    To add the target attribute, target style scale is sampled from a distribution mapped for the given attribute ($j$), e.g. blonde, brown or encoded and projected from a reference image. }
    \label{fig:generator}
\end{figure}

For image to image translation task, we set an encoder-decoder based architecture and latent space translation in the middle as given in Fig. \ref{fig:generator}.
We perform the translation in the encoded latent space, $e$, which is obtained by $e = E(x)$ where $E$ refers to the encoder. The encoded features go through a transformation $T$ which is discussed in the next section.
The transformed features are then decoded by $G$ to reconstruct the translated images.
The image generation pipeline following feature encoding is described in Eq. \ref{eqn:decoder}.
 \begin{align}
     e' = T(e, \alpha, i) \nonumber \\
     x' = G(e') \label{eqn:decoder}
 \end{align}
 
Previous image-to-image translation networks \cite{li2021image,yang2021l2m,starganv2} set a shallow encoder decoder architecture to translate an image and a separate deep network for style encoding.
In most cases, the style encoder includes separate branches for each tag.
The shallow architecture that is used to translate images prevents the model from making drastic changes in the images and this helps preserving the identity of the persons.
Our framework is different as we do not employ a separate style encoder and instead have a deep encoder-decoder architecture for translation.
That is because to be able to organize the latent space in an interpretable way, our framework requires a full understanding of the image and therefore a larger receptive field; deeper network architecture.
A deep architecture with decreasing size of feature size, on the other hand, faces the challenges of reconstructing all the fine details from the input image.

With the motivation of helping the network to preserve tag independent features such as the fine details from background, we use skip connections between our encoder and decoder.
However, we observe that the flow of information should be limited to force the encoder-decoder architecture learn facial attributes and well-organized latent representations. 
Because of that reason, we only allow skip connection at low resolution.
This design is extensively justified in our Ablation Studies.



\subsection{Translation Module} \label{style-shift}
To achieve a style transformation, we perform the tag-based feature manipulation in a linear fashion in the latent space.
First, we set a feature direction matrix $A$ which contains learnable feature directions for each tag.
In our formulation $A_i$ denotes the learned feature direction for tag $i$.
Direction matrix $A$ is randomly initialized and learned during the training process.

Our translation module is formulated in Eq. \ref{eqn:translator}, which adds the desired shift on top of the encoded features $e$ similar to \cite{voynov2020unsupervised}.

\begin{equation}
T(e, \alpha, i) = e + \alpha \times A_i
\label{eqn:translator}
\end{equation}

We compute the shift by subtracting target style from the source style as given in Eq \ref{eqn:alphat}.

\begin{equation}
\alpha = \alpha_t - \alpha_s
\label{eqn:alphat}
\end{equation}

Since the attributes are designed as linear steps in the learnable directions, we find the style shift by subtracting the target attribute scale from source attribute scale.
This way the same target attribute $\alpha_t$ can have the same impact on the translated images no matter what the attributes were of the original images.
For example, if our target scale corresponds to brown hair, the source scale can be coming from an image with blonde or back hair but since we take a step for difference of the scales, they can be both translated to an image with the same shade of brown hair.

To extract the target shifting scale for feature (tag) $i$, $\alpha_t$, there are two alternative pathways.
The first pathway, named as latent-guided path, samples a $z \in \mathcal{U}[0,1)$ and applies a linear transformation $\alpha_t = w_{i,j} \cdot z + b_{i,j}$, where $\alpha_t$ denotes sampled shifting scale for tag $i$ and attribute $j$.
Here tag $i$ can be hair color and attribute $j$ can be blonde, brown, or back hair. For each attribute we learn a different transformation module which is denoted  as $M_{i,j}(z)$.
Since we learn a single direction for every tag for example for hair color, this transformation module can put the initially sampled $z$'s into correct scale in the linear line based on the target hair color attribute.
As the other alternative pathway, we encode the scalar value $\alpha_t$ in a reference-guided manner.
We extract $\alpha_t$ for tag $i$ from a provided reference image by first encoding it into the latent space, $e_r$, and projecting $e_r$ via by $A_i$ as given in Eq. \ref{eqn:proj}.

\begin{equation}
\alpha_t = P(e_r, A_i) =  \dfrac{e_r \cdot A_i}{||A_i||}
\label{eqn:proj}
\end{equation}

In the reference guidance set-up, we do not use the information of attribute $j$, since it is encoded by the tag $i$ features of the image.

The source scale, $\alpha_s$, is obtained by the same way we obtain  $\alpha_t$ from reference image.
We perform the projection for the corresponding tag we want to manipulate, $i$, by $P(e, A_i)$.
We formulate our framework with the intuition that the scale controls the  amount of feature to be added.
Therefore, especially when the attribute is copied over from a reference image, the amount of features that will be added will be different based on the source image.
It is for this reason, we find the amount of shift by subtraction as given in Eq. \ref{eqn:alphat}.
Our framework is intuitive and relies on a single encoder-decoder architecture.
Fig. \ref{fig:generator} shows the overall pipeline.

\subsection{Training pathways}

Modifying the translation paths defined by \cite{li2021image}, we train our network using two different paths. For each iteration to optimize our model, we sample a tag $i$ for shift direction, a source attribute $j$ as the current attribute and a target attribute $\hat{j}$.

\textbf{Non-translation path.}
To ensure that the encoder-decoder structure preserves details of the images, we perform a reconstruction of the input image without applying any style shifts.
The resulting image is denoted as $x_n$ as given in Eq. \ref{eqn:non-tr}. 
\begin{equation}
    x_n = G(E(x))
    \label{eqn:non-tr}
\end{equation}


\begin{figure}[t]
    \centering
    \includegraphics[width=1.0\textwidth]{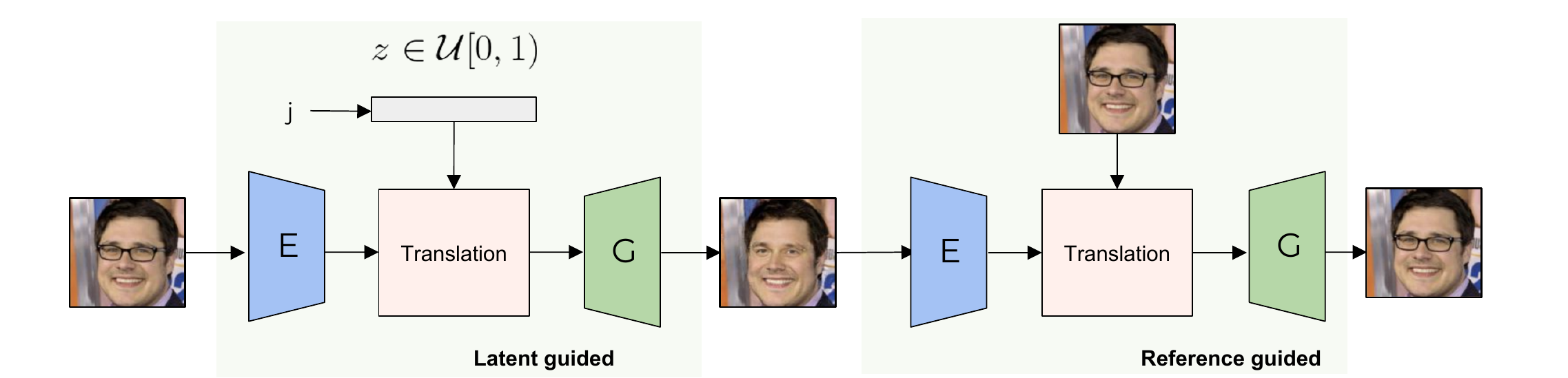}
    \caption{Overview of cycle translation path.}
    \label{fig:cycle}
\end{figure}

\textbf{Cycle-translation path.}
We apply a cyclic translation to ensure that we get a reversible translation from a latent guided scale.
In this path, as shown in Fig. \ref{fig:cycle}, we first apply a style shift by sampling $z \in \mathcal{U}[0,1)$ and obtaining target $\alpha_t$ with $M_{i,\hat{j}}(z)$ for target attribute $\hat{j}$.
The translation uses $\alpha$ that is obtained by subtracting $\alpha_t$ from the source style.
Decoder generates an image, $x_t$ as given in Eq. \ref{eqn:cycle-t1}  where $e$ is encoded features from input image $x$, $e=E(x)$.
$x_t$ refers to the image without glasses in Fig.  \ref{fig:cycle}.

\begin{align}
    x_t = G(T(e, M_{i,j}(z) - P(e, i), i))\label{eqn:cycle-t1}
\end{align}

Then by using the original image, $x$, as a reference image, we aim to reconstruct the original image by translating $x_t$.
Overall, this path attempts to reverse a latent-guided style shift with a reference-guided shift.
The second translation is given in Eq. \ref{eqn:cycle-tr} where $e_t=E(x_t)$.

\begin{align}
    x_c = G(T(e_t, P(e, i) - P(e_t, i), i))\label{eqn:cycle-tr}
\end{align}

In our learning objectives, we use $x_n$ and $x_c$ for reconstruction and $x_t$ and $x_c$ for adversarial losses, and $M_{i,j}(z)$ for the shift reconstruction loss. Details about the learning objectives are given in the next section.


\subsection{Learning objectives}
Given an input image $x_{i,j} \in \mathcal{X}_{i,j}$, where $i$ is the tag to manipulate and $j$ is the current attribute of the image, we optimize our model with the following objectives. In our equations, $x_{i,j}$ is shown as $x$.

\textbf{Adversarial Objective.}
During training, our generator performs a style-shift either in a latent-guided way or a reference-guided way, which results in a translated image. 
In our adversarial loss, we receive feedback from the two steps of cycle-translation path.
As the first component of the adversarial loss, we feed a real image $x$ with tag $i$ and attribute $j$ to the discriminator as the real example.
To give adversarial feedback to latent-guided path, we use the intermediate image generated in cycle-translation path, $x_t$.
Finally, to provide adversarial feedback to reference-guided path, we use the final outcome of the cycle-translation path $x_c$.
Only $x$ acts as real image, both $x_t$ and $x_c$ are translated images, and they are treated as fake images with different attributes.
The discriminator aims at classifying whether an image, given its tag and attribute, is real or not. 
The objective is given in Eq \ref{eqn:adv}.

\begin{equation}
\begin{split}
    \mathcal{L}_{adv} = 2log(D_{i,j}(x)) 
    +  log(1 - D_{i, \hat{j}} (x_t))
    +  log(1 - D_{i, j} (x_c))
    \label{eqn:adv}
\end{split}
\end{equation}



\textbf{Shift Reconstruction Objective.}
As the cycle-consistency loss performs reference-guided generation followed by latent-guided generation, we utilize a loss function to make these two methods consistent with each other \cite{lee2018diverse,huang2018multimodal,li2019attribute,li2021image}.
Specifically, we would like to obtain the same target scale, $\alpha_t$, both from the mapping and from the encoded reference image generated by the mapped $\alpha_t$. The loss function is given in Eq. \ref{eqn:l_shift}.
\begin{equation}
    \mathcal{L}_{shift} = ||M_{i,j}(z) - P(e_t, i)||_1
    \label{eqn:l_shift}
\end{equation}
Those parameters, $M_{i,j}(z)$ and $P(e_t, i)$, are calculated for the cycle-translation path as given in Eq. \ref{eqn:cycle-t1} and \ref{eqn:cycle-tr}.

\textbf{Image Reconstruction Objective.}
In all of our training paths, the purpose it to be able to re-generate the original image again.
To supervise this desired behavior, we use $L_1$ loss for reconstruction loss.
In our formulation $x_n$ and $x_c$ are outputs of non-translation path and cycle-translation path, respectively. Formulation of this objective is provided in Eq. \ref{eqn:rec_loss}.

\begin{equation}
    \begin{split}
        \mathcal{L}_{rec} = ||x_n - x||_1 + 
        ||x_c - x||_1
    \end{split}
    \label{eqn:rec_loss}
\end{equation}

\textbf{Orthogonality Objective.}
 To encourage the orthogonality between directions, we use soft orthogonality regularization based on Frobenius norm, which is given in Eq. \ref{eqn:ortho_loss}.
 This orthogonality further encourages a disentanglement in the learned  style directions.
 
\begin{equation}
\mathcal{L}_{ortho} =  {\|A^{T}A - I\|_F} 
  \label{eqn:ortho_loss}
\end{equation}

\textbf{Full Objective.}
Combining all of the loss components described, we reach to the overall objective for optimization as given in Eq. \ref{eqn:full_loss}.
We additionally add L1 loss on the matrix $A$ parameters to encourage its sparsity.

\begin{equation}
    \begin{split}
        \underset{E,G,M,A}{\min} \underset{D}{\max} \lambda_{a}\mathcal{L}_{adv} +  \lambda_{s} \mathcal{L}_{shift} + \lambda_{r} \mathcal{L}_{rec}+
        \lambda_{o} \mathcal{L}_{ortho}+
        \lambda_{sp} \mathcal{L}_{sparse}
    \end{split}
    \label{eqn:full_loss}
\end{equation}

To control the dominance of each loss component, we use $\lambda_{a}, \lambda_{s}$, $\lambda_{r}$,  $\lambda_{o}$, and $\lambda_{sp}$ hyperparameters. 
These hyperparameter values and training details are given in Supplementary.

\section{Experiments}
\label{sec:exp}

\subsection{Dataset and Settings}

We train our model on CelebA-HQ dataset \cite{celeba} which contains 30,000 face images.
To extensively compare with state-of-the-arts, we follow two training-evaluation protocols as follows:

\textbf{Setting A.} In our first setting, we follow the set-up from HiSD \cite{li2021image}. Following HiSD, we use the first 3000 images of CelebA-HQ dataset as the test set and 27000 as the training set.
These images include annotations for different attributes from which we use hair color, presence of glass, and bangs attributes for translation task in this setting.
Hair color attribute includes 3 tags, black, brown, and blonde whereas the other attributes are binary.
The images are resized to $128\times128$.
Following the  evaluation protocol proposed by HiSD \cite{li2021image}, we compute FID scores on bangs addition task.
For each test image without bangs, we translate them to images with bangs with latent and reference guidance.
In latent guidance, 5 images are generated for each test image by randomly sampling scale from a uniform distribution.
Then this generated set of images are compared with images that have attribute bangs in terms of their FIDs.
FIDs are calculated for these 5 sets and averaged. 
For reference guidance, we randomly pick 5 references images to extract the style scale.
FIDs are calculated for these 5 sets separately and averaged.

\textbf{Setting B.} In this setting, we follow the set-up from L2M-GAN \cite{yang2021l2m}. The training/test split is obtained by re-indexing each image in CelebA-HQ back to the original CelebA and following the standard split of CelebA. This results in 27,176 training and 2,824 test images.
Models are trained for hair color, presence of glasses,  bangs, age, smiling, and gender attributes.
Images are resized to $256\times256$ resolution.
For evaluation, smiling attribute is used following L2M-GAN \cite{yang2021l2m}.
It is noted that smiling is one of the most challenging among the CelebA facial attributes because adding/removing
a smile requires high-level understanding of the input face
image for modifying multiple facial components simultaneously.
FIDs are calculated for adding and removing the smile attribute.

\subsection{Results} We extensively compare our results with other competing methods in Table \ref{table:results_all}. 
In Setting A, as given in Table \ref{table:results_setA}, we compare with SDIT \cite{wang2019sdit}, StarGANv2 \cite{starganv2}, Elegant \cite{xiao2018elegant}, and HiSD \cite{li2021image} models.
Among these methods, HiSD learns a hierarchical style disentanglement whereas StarGANv2 learns a mixed style code.
Therefore, StarGANv2 when translating images also does other unnecessary manipulations and does not strictly preserve the identity.
Our work is most similar to HiSD as we also learn disentangled style directions.
However, HiSD learns feature based local translators which is an approach known to be successful on local edits, e.g. bangs.
Ours results show that VecGAN achieves significantly better quantitative results than HiSD both in latent guided and reference guided evaluations even though they are compared on a local edit task.

\begin{table}[t]
\begin{subtable}[t]{0.45\textwidth}
    \centering
    \begin{tabular}{|l|l|l|}
\hline
\textbf{Method} &  \textbf{Lat.} & \textbf{Ref.} \\
\hline
SDIT \cite{wang2019sdit} & 33.73 & 33.12 \\
StarGANv2 \cite{starganv2} & 26.04 & 25.49 \\
Elegant \cite{xiao2018elegant} & - & 22.96   \\ 
HiSD \cite{li2021image} &  21.37 &  21.49 \\
\hline
VecGAN (Ours) &  \textbf{20.17} & \textbf{20.72} \\
\hline
\end{tabular}
    \caption{Quantitative results for Setting A.  Lat: Latent guided, Ref: Reference guided.
    FID scores are given. Lower is better.}
    \label{table:results_setA}
    \end{subtable}
    \hspace{\fill}
    \begin{subtable}[t]{0.53\textwidth}
    \centering
    \begin{tabular}{|l|l|l|l|}
\hline
\textbf{Method} &  \textbf{FID (+)} & \textbf{FID (-)} & \textbf{Avg}\\
\hline
StarGAN \cite{choi2018stargan} & 32.6 & 38.6 & 35.6 \\
CycleGAN \cite{zhu2017unpaired}  & 22.5 & 24.4 & 23.5 \\
Elegant \cite{xiao2018elegant} & 39.7 & 42.9  & 41.3 \\ 
PA-GAN \cite{he2020pa} & 20.5 & 21.4 & 21.0\\
InterFaceGAN \cite{shen2020interpreting} & 24.8 & 24.9 & 24.9 \\
L2M-GAN \cite{yang2021l2m} & 17.9 & 23.3 & 20.6  \\
\hline
VecGAN (Ours) & \textbf{17.7} & \textbf{20.3} & \textbf{19.0} \\
\hline
\end{tabular}
    \caption{Quantitative results for Setting B. FID (+) (or FID (-)) denotes the FID score for adding (or removing) a smile. }
    \label{table:results_setB}
    \end{subtable}
    \caption{Comparisons with state-of-the-art competing methods. Please refer to Section \ref{sec:exp} for details on training and evaluation protocol of Setting A and B.}
    \label{table:results_all}
\end{table}


\begin{table}[t]
\centering
\begin{tabular}{|l|l|l|l|l|l|l|}
\hline
& \multicolumn{2}{c}{\textbf{Smiling (+)}} & \multicolumn{2}{c}{\textbf{Smiling (-)}} & \multicolumn{2}{c|}{\textbf{Smiling (Avg)}} \\
\hline
\textbf{Comparisons}  & Quality & Fidelity & Quality & Fidelity & Quality & Fidelity \\
\hline
VecGAN (Ours) vs L2M-GAN  & 57.96\% & 70.94\% & 60.93\% & 77.50\% & 59.45\% & 74.22\% \\
VecGAN (Ours) vs InterFaceGAN & 88.13\%  & 91.56\% & 77.50\% & 90.62\%  & 82.82\% & 91.09\% \\
\hline
\end{tabular}
\caption{User study results conducted with smiling attribute.
Smiling (+) denotes the results of adding a smile, Smiling (-) refers to the results of removing a smile, and Smiling (avg) denotes the average of Smiling (+) and Smiling (-). Percentages show the preference rates of our method versus the other competing method.}
\label{tab:user}
\end{table}

\begin{figure}[t]
    \centering
    \includegraphics[width=1.0\textwidth]{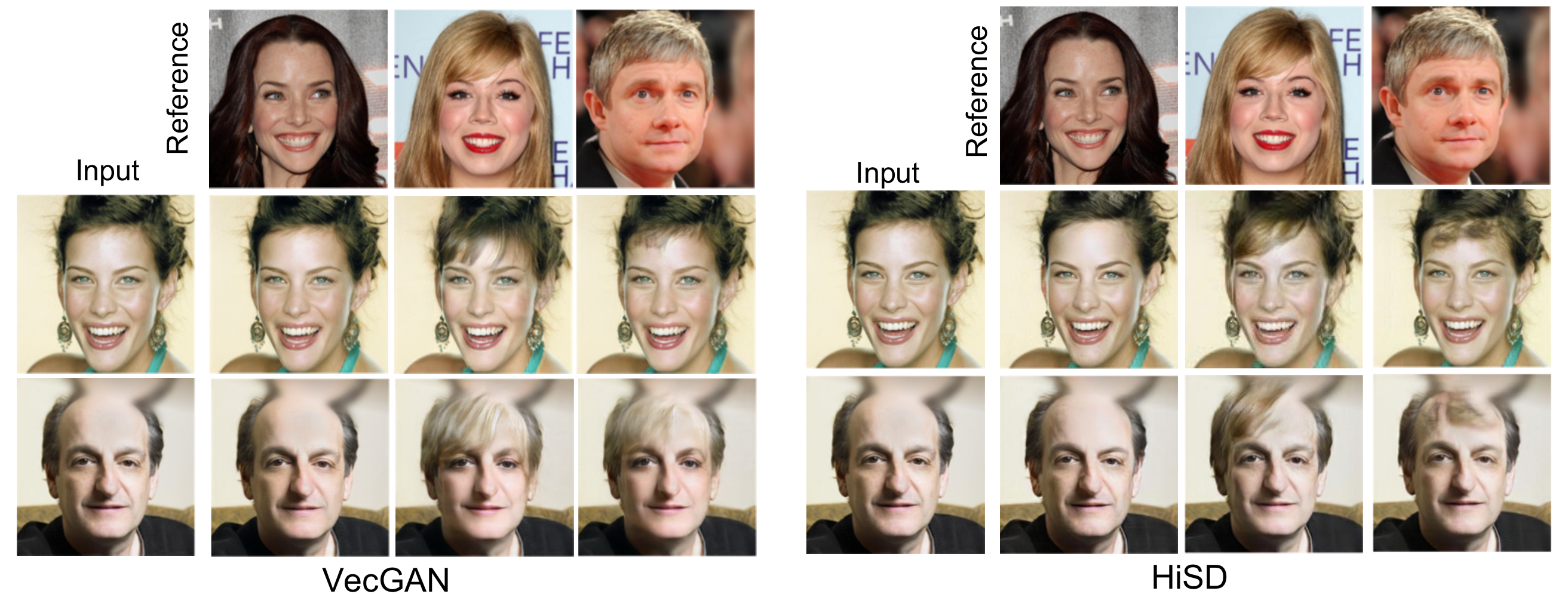}
    \caption{Qualitative results of bangs attribute of our model (VecGAN) and HiSD. 
    In the second example, we provide a very challenging sample  where VecGAN even though not perfect achieves significantly better results than HiSD.}
    \label{fig:hisd_comp}
\end{figure}

\begin{figure}[h]
    \centering
    \includegraphics[width=1.0\textwidth]{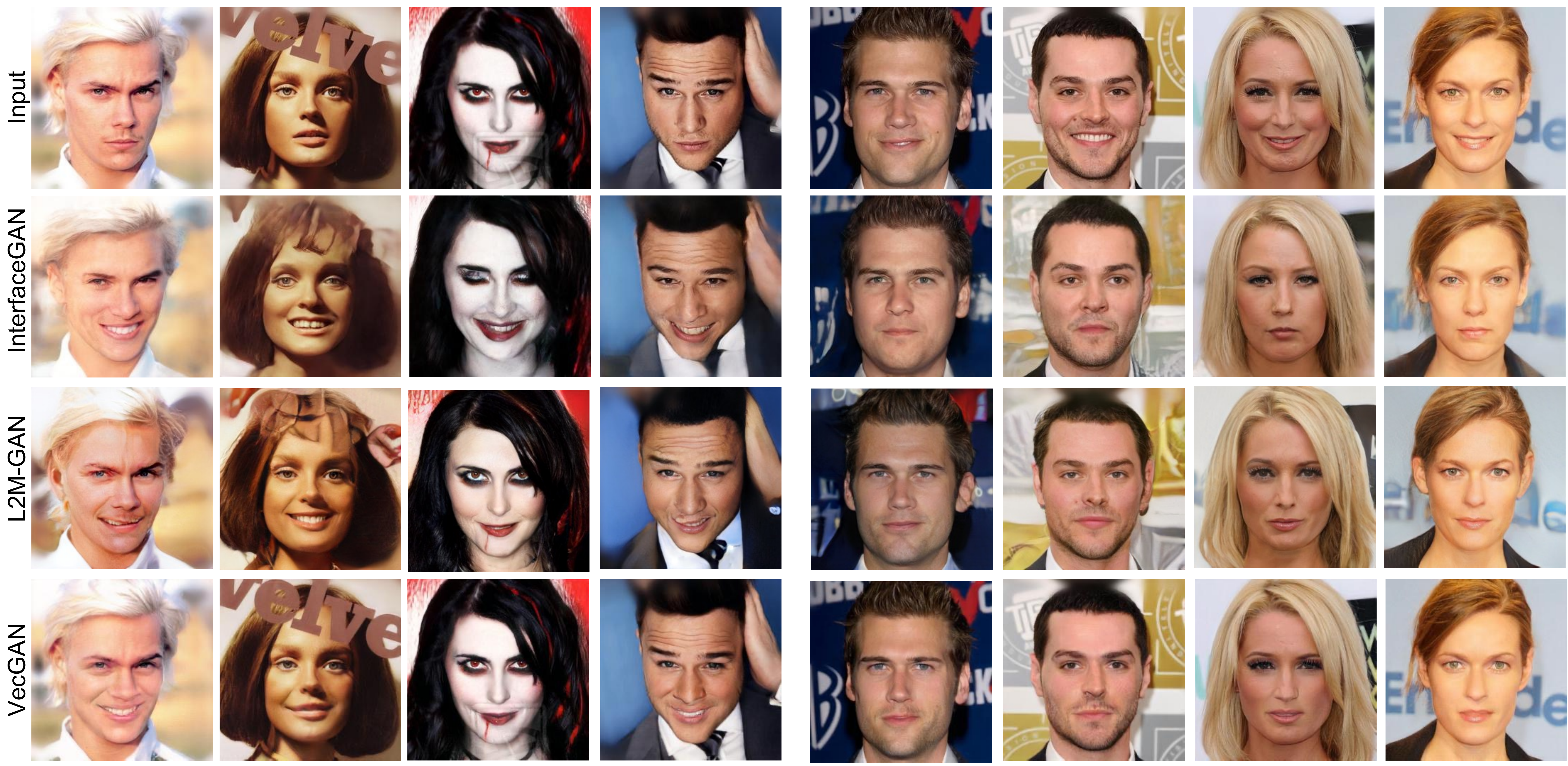}
    \caption{Qualitative results of smile attribute of our model (VecGAN), L2M-GAN, and InterFaceGAN. The first four examples show smile addition and the other four shows smile removal manipulations.}
    \label{fig:l2mgan_comp}
\end{figure}

Fig. \ref{fig:hisd_comp} shows reference guided results of our model versus HiSD.
We compare with HiSD since it provides with the best results after ours.
As can be seen from Fig. \ref{fig:hisd_comp}, both methods achieve attribute disentanglement, they do not change any other attribute of the image than the bangs tag.
However, HiSD outputs artifacts especially for the reference image from the last column. On the other hand, VecGAN outputs higher quality results.
As the second example, we pick a very challenging example to compare these methods.
Even though, our results can be further improved to look more realistic, it achieves significantly better outputs with no artifacts compared to HiSD.

In our second set-up of evaluation, we compare our method with many state-of-the-art methods as given in Table \ref{table:results_setB}.
We compare with StarGAN \cite{choi2018stargan}, CycleGAN \cite{zhu2017unpaired}, Elegant \cite{xiao2018elegant}, PA-GAN \cite{he2020pa}, InterFaceGAN \cite{shen2020interpreting}, and L2M-GAN \cite{yang2021l2m}.
For InterFaceGAN, we use the GAN Inversion \cite{zhu2020domain} as the encoder and pretrained StyleGAN \cite{karras2019style} as the generator backbone.
As can be seen from Table \ref{table:results_setB}, we achieve significantly better scores on both settings and in average.

In our visual comparisons, we mainly focus on  L2M-GAN and InterFaceGAN since L2M-GAN is the second best model after ours and InterFaceGAN shares the same intuition with our model and performs edits by latent code manipulation. 
The results are shown in Fig. \ref{fig:l2mgan_comp} where the first four examples show smile addition and the other four examples show smile removal manipulations.
The most prominent limitation of L2M-GAN and InterFaceGAN is that they do not preserve the other attributes of images, especially on the background whereas VecGAN does a very good job at that.
Smile attribute addition and removal of L2M-GAN is better than InterFaceGAN, however, worse than ours. 
VecGAN is the only method among them that can produce manipulated images with high fidelity to the originals with only targeted attribute manipulated in a natural and realistic way.

We also conduct a user study on the first 64 images of validation set among 10 users. We set an A/B test and provide users with input images and translations obtained by VecGAN and other competing methods.
The left-right order is randomized to ensure fair comparisons.
We perform two separate tests. 1) Quality: We ask users to select the best result according to i) whether the smile attribute is correctly added, ii) 
whether irrelevant facial attributes preserved, iii) and overall whether the output image looks realistic and high quality.
2) Fidelity: We ask users to pay attention if details from the input image is preserved in addition to the quality.
When only asked for quality, users pay attention to facial attributes and do not pay much attention to the background, ornament, details of hair of the image, and so on.
In this test, we remind the users to pay attention to those as well.
Table~\ref{tab:user} shows the results of the user study.
Users preferred our method as opposed to L2M-GAN $59.45\%$ of the time ($50\%$ is tie), and as opposed to InterFaceGAN $82.82\%$ of the time for the quality measure in average of smile addition and removal results.
When users asked to pay attention to non-facial attributes as well, they preferred our method as opposed to L2M-GAN $74.22\%$ of the time, and as opposed to InterFaceGAN $91.09\%$ of the time in average.

\subsection{Ablation Study} 

\begin{figure}[t]
\begin{floatrow}
\ffigbox{%
 {\includegraphics[width=0.4\textwidth]{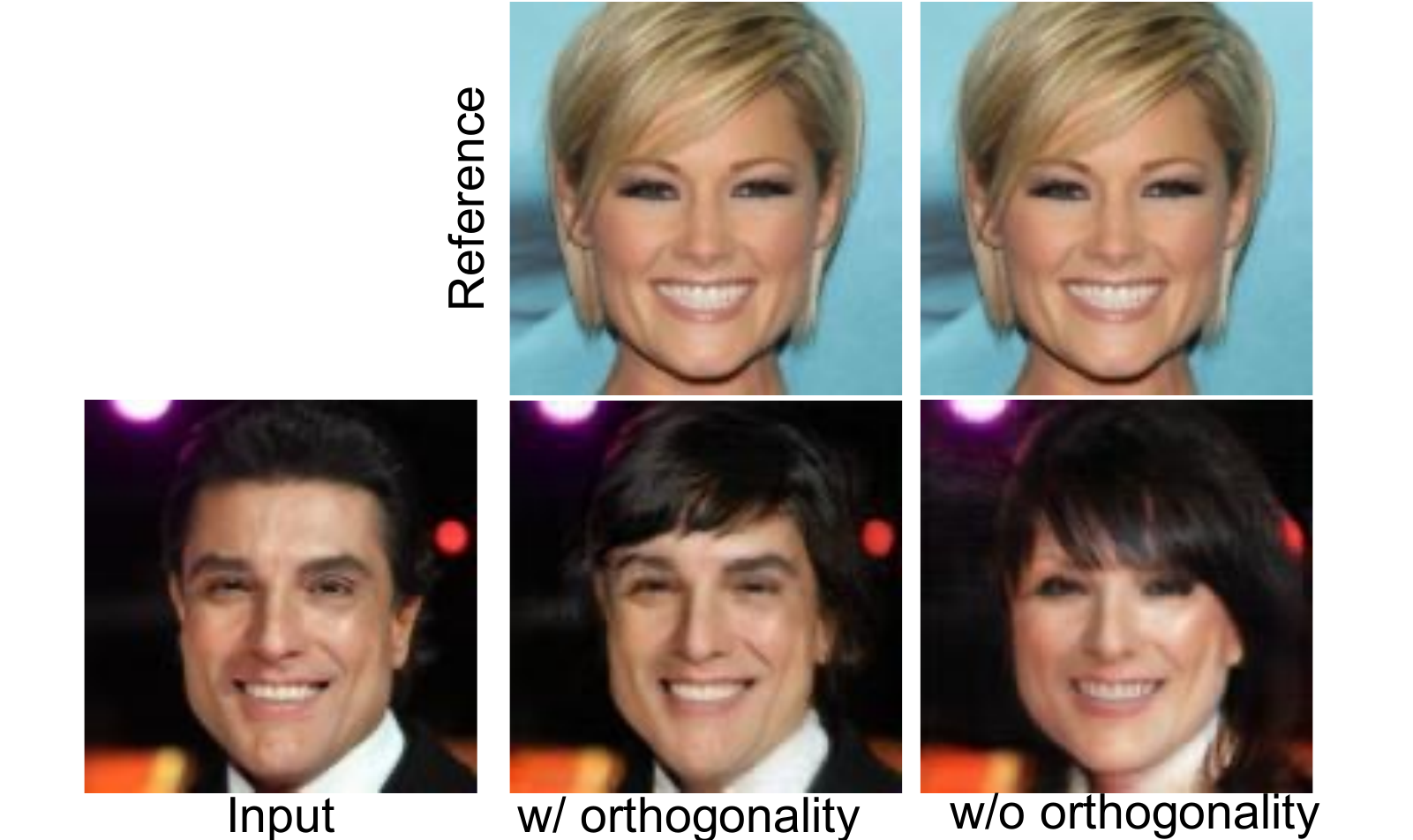}}
}{%
  \caption{Qualitative results of ablation study of orthogonality loss. Bangs tag transferred from the reference image.}%
  \label{fig:abl}
}
\capbtabbox{%
    \begin{tabular}{|l|l|l|}
\hline
\textbf{Method} &  \textbf{Lat.} & \textbf{Ref.} \\ \hline
Shallow &  21.30 & 20.94 \\
Deep w/o skip & 88.62 &  127.65 \\
Deep all skip  & 273.80 & 273.97 \\
Ours & \textbf{20.17} & \textbf{20.72} \\
\hline
w/o Orthogonality & 21.98 & 22.50 \\
w/o Sparsity & 24.07 & 22.43\\
\hline
\end{tabular}
}{%
  \caption{FID results of ablation study with Setting A. Lat: Latent guided, Ref: Reference guided.}%
  \label{tab:abl}
  
}

\end{floatrow}
\end{figure}

We conduct ablation studies for  network architecture and loss objectives as given in Table \ref{tab:abl}.
We first experiment with a shallower architecture where encoder decreases the input dimension of $128\times128$ to a spatial dimension of $8\times8$.
This version gives reasonable scores, however, we are interested in a better latent space organization.
For that, we use a deeper encoder-decoder architecture where encoded latent space goes as low as $1\times1$ which we refer as deep architecture.
Deep architecture without skip connections is not able to minimize the reconstruction objective and results in a high FID.
On the other hand, deep architecture with a skip connection at each resolution from encoder to decoder can minimize the reconstruction loss however the latent space is not well organized since the model tends to pass all the information from the encoder which instabilizes the training.
Our architecture with single skip layer at resolution $32\times32$ provides a good balance between the information flow from encoder-decoder and the latent space bottleneck.

Next, we experiment the effect of loss functions. First, we remove the orthogonality loss of $A$ directions.
This results in worse FID scores but more importantly we observe that the styles are not disentangled, e.g. changing bangs attribute changes the gender as can be seen in Fig. \ref{fig:abl}. 
Even without this loss function, we observe that during training the orthogonality loss of $A$ decreases but to a higher value than when this loss is added to the final objective. That is because the framework and other loss objectives also encourage the disentanglement of attribute manipulations and it shows in the orthogonality of direction vectors.
This also shows the importance of orthogonality in style disentanglement and this targeted loss helps improve that significantly.
We also observe that sparsity loss applied on the directional vectors stabilizes the training and without that FIDs are much higher.

\begin{figure}[t]
    \centering
    \includegraphics[width=1.0\textwidth]{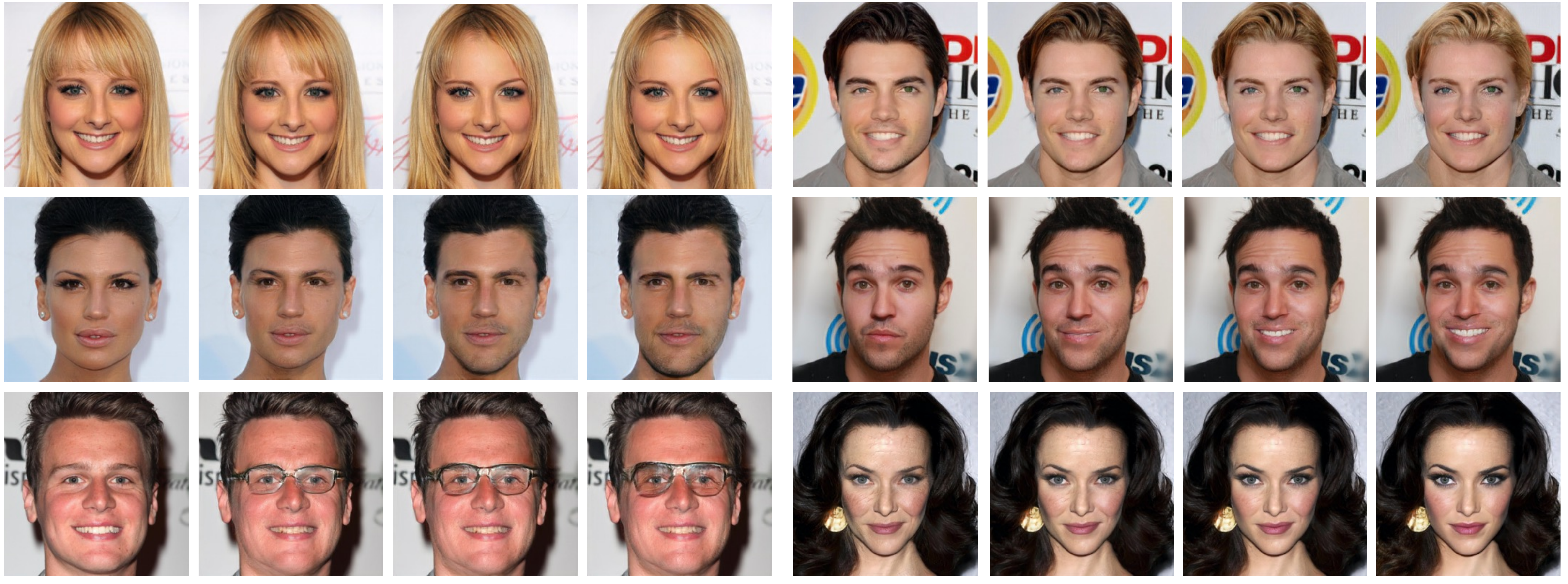}
    \caption{Results of changing the strength of a manipulation gradually. Each example shows a different attribute manipulation.
    Rows show bangs, hair color, gender, smile, glasses, and age manipulations in this order.}
    \label{fig:alpha}
\end{figure}

\subsection{Other Capabilities of VecGAN} 
\textbf{Gradually Increased Scale.} 
We translate images with gradually increased attribute strength as shown in Fig. \ref{fig:alpha}.
We plot the manipulation results on six different attributes.
These results show that attributes that are designed as linear transformations are disentangled, and changing one attribute does not affect the other components.
In these results, as scales are gradually increased, the strength of the tag smoothly increases with the identity of the person preserved.

\begin{figure}[t]
   \begin{subfigure}[b]{0.45\textwidth}
    \centering
    \includegraphics[width=0.95\textwidth]{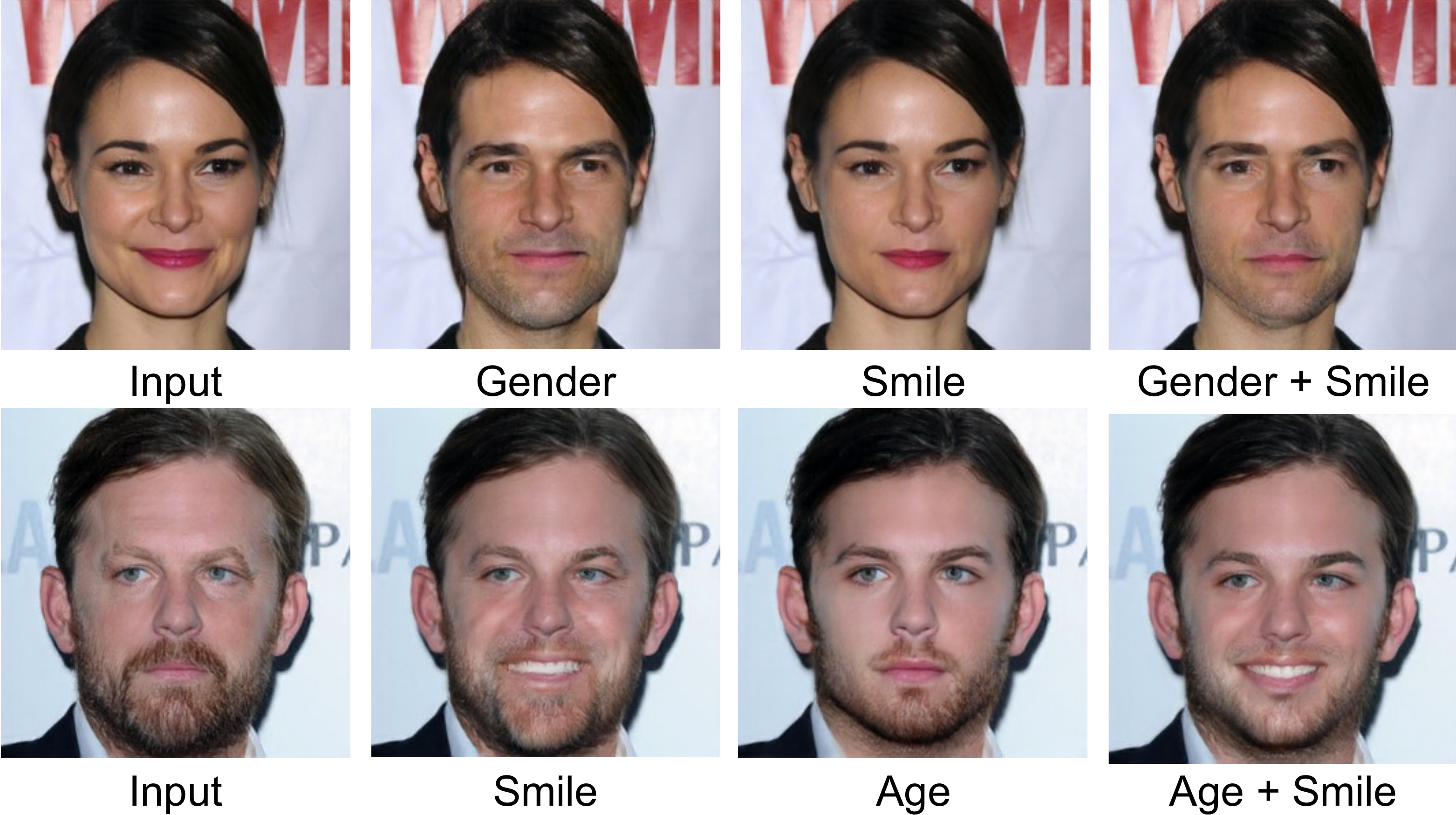}
      \caption{Multi-attribute editing results.}
    \label{fig:multi_comp}
\end{subfigure}
    \begin{subfigure}[b]{0.50\textwidth}
    \centering
    \includegraphics[width=1.0\textwidth]{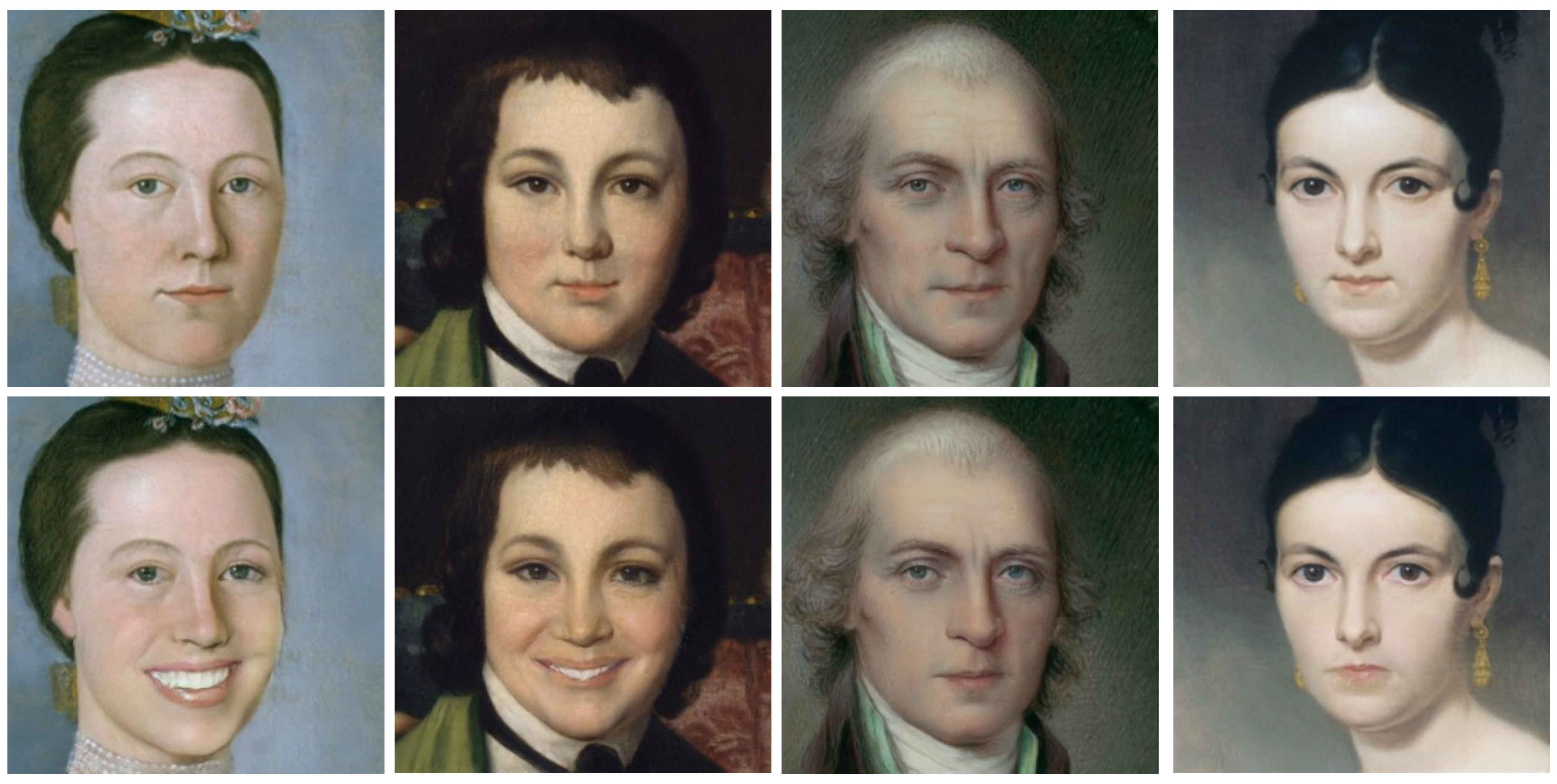}
    \caption{Generalization results.}
      \label{fig:multi_gen}
\end{subfigure}   
    \caption{Results of multi-attribute editing and cross-dataset generalization results of VecGAN.}

\end{figure}

\textbf{Multi-tag Edits.} We additionally experiment with multi-tag manipulation.
To change two attributes, instead of encoding and decoding the image twice with a translation in between each time, we perform two translation operations in the latent code simultaneously.
That is we apply Eq. \ref{eqn:translator} twice for two different $i$.
Fig. \ref{fig:multi_comp}  shows results of the multi-tag edits. 
In the first row, we consider gender and smile tags, and first edit those attributes individually. In the last coloumn, we edit the image with these two tags simultaneously.
The second row shows a similar experiment with smile and age tags. We observe that VecGAN provides with disentangled tag control and can successfully edit  tags independently.

\textbf{Generalization to other domains.} We apply VecGAN model to MetFace dataset \cite{karras2020training} without any retraining.
The results are provided in Fig. \ref{fig:multi_gen}. The first row shows source images, and the second row shows outputs of our model.
In the first two examples, we increase the smile attribute, and in the other two, we decrease it. 
The results show that VecGAN has a good generalization ability and works reasonably well across datasets.



\section{Conclusion}

This paper introduces VecGAN, an image-to-image translation framework with interpretable latent directions.
This framework includes a deep encoder and decoder architecture with latent space manipulation in between.
Latent space manipulation is designed as vector arithmetic where for each attribute, a linear direction is learned.
This design is encouraged by the finding that well-trained generative models organize their latent space as disentangled representations with meaningful directions in a completely unsupervised way.
Each change in the architecture and loss functions is extensively studied and compared with state-of-the-arts. Experiments show the effectiveness of our framework.

\appendix

\section{More comparisons}

\begin{figure}[h]
    \centering
    \includegraphics[width=1.0\textwidth]{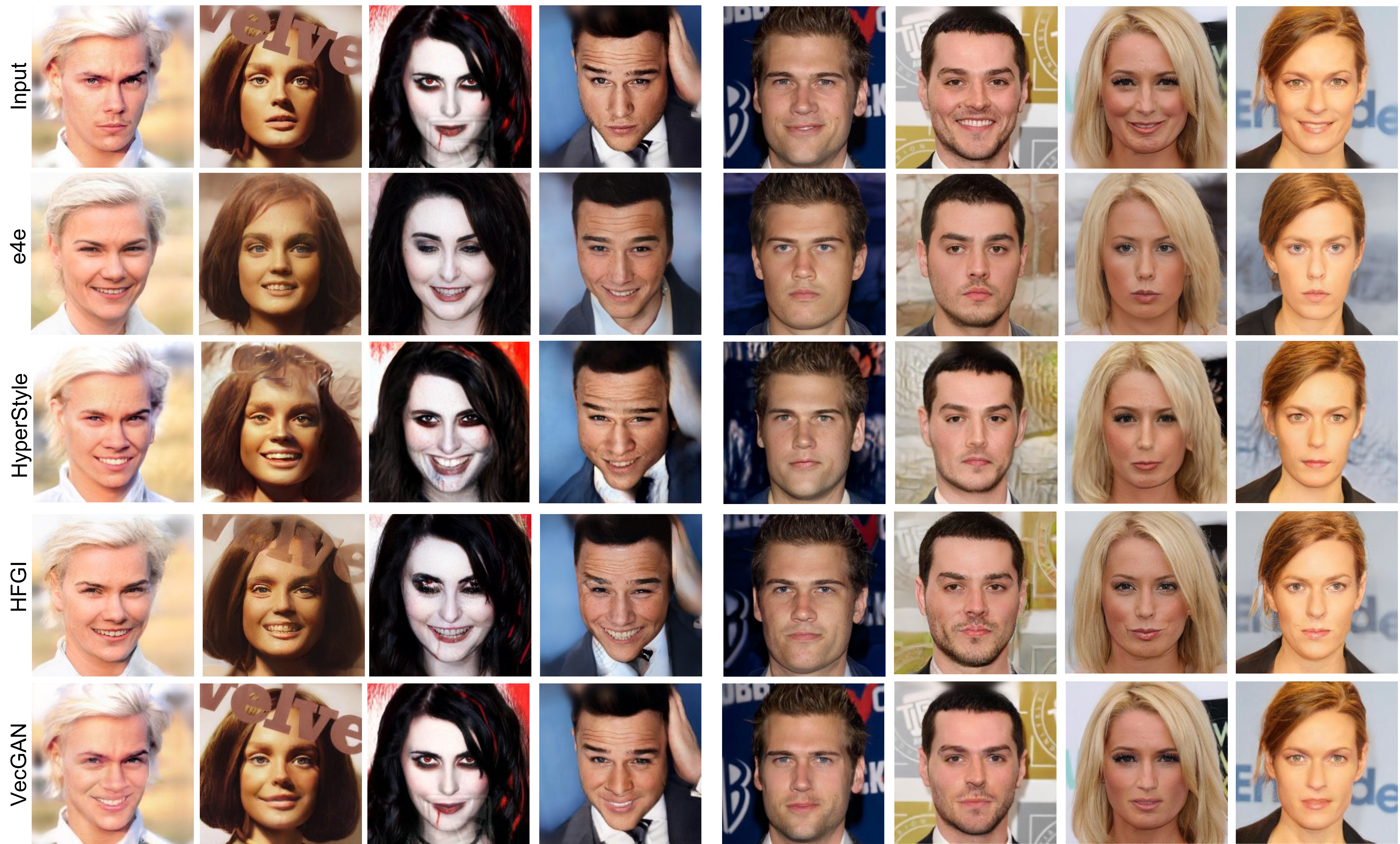}
    \caption{Qualitative results of smile attribute of our model (VecGAN) and other StyleGAN based models. }
    \label{fig:sup_l2mgan_comp}
\end{figure}

In Fig. \ref{fig:sup_l2mgan_comp}, we compare our method with other methods that are proposed to invert images to StyleGANv2 space and perform edits via the pretrained StyleGANv2.
We compare with e4e \cite{tov2021designing}, HyperStyle \cite{alaluf2022hyperstyle}, and HFGI \cite{wang2022high}.
Same input examples are used from Fig. \ref{fig:l2mgan_comp} main paper.
e4e as also stated in their paper outputs results with worse distortion (input-output similarity) but better edits.
HyperStyle and HFGI are concurrent works with improved fidelity to the input image but still significantly worse than our method both in edit quality and reconstruction quality of the input details.

\begin{figure}
    \centering
    \includegraphics[width=1.0\linewidth]{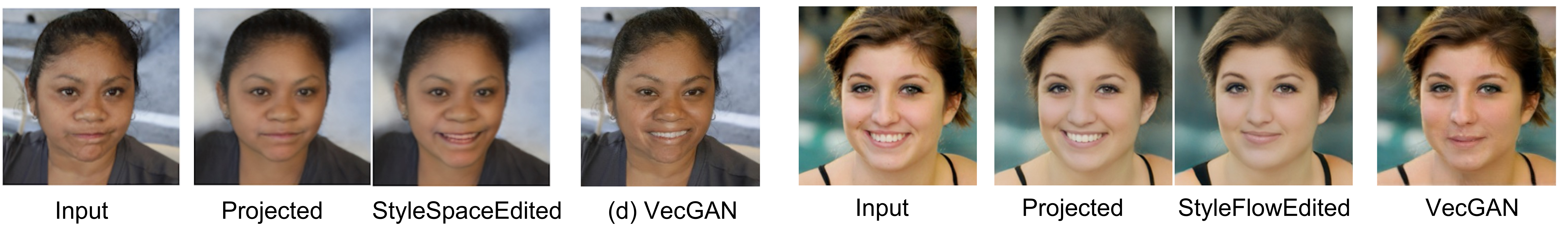}
    \caption{ Qualitative results of smile attribute of our model (VecGAN) and other
StyleGAN based editing models.}
    \label{fig:sup_stylespace}
\end{figure}

We additionally compare with StyleFlow \cite{abdal2021styleflow} and StyleSpace \cite{wu2021stylespace} in Fig. \ref{fig:sup_stylespace}.
For both examples, we take their real image editing example from their papers and feed the input crops to VecGAN for comparison.
As can be seen from Fig. \ref{fig:sup_stylespace}, both methods suffer from the limitations of the projection method as inputs are not faithfully reconstructed.  Additionally, the edit is not perfectly disentangled in StyleFlow example as the strap of the top changes when smile is modified.
VecGAN achieves significantly better results in these examples.

\section{Additional Quantitative Results}

\begin{table}[]
    \centering
    \begin{tabular}{|l|c|c|c|}
        \hline
        Method & KID(+) & KID(-) & KID (Avg)\\
        \hline
        L2M-GAN & 0.01010 & 0.00942 &  0.00976\\
        InterfaceGAN & 0.00603 & 0.00671 & 0.00637 \\
        VecGAN &  \textbf{0.00188} & \textbf{0.00328} & \textbf{0.00258}\\
        \hline
    \end{tabular}
     \caption{Quantitative results for Setting B - Smile attribute.}
    \label{tab:reb_kid}
\end{table}

In Table \ref{tab:reb_kid}, we compare VecGAN and other competing methods with KID metric \cite{binkowski2018demystifying}.
Same as in FID evaluation, VecGAN achieves significantly better results.

\section{Model Architecture}
In this section, we provide architectural details of VecGAN.

\begin{figure}
    \centering
    \includegraphics[width=1\linewidth]{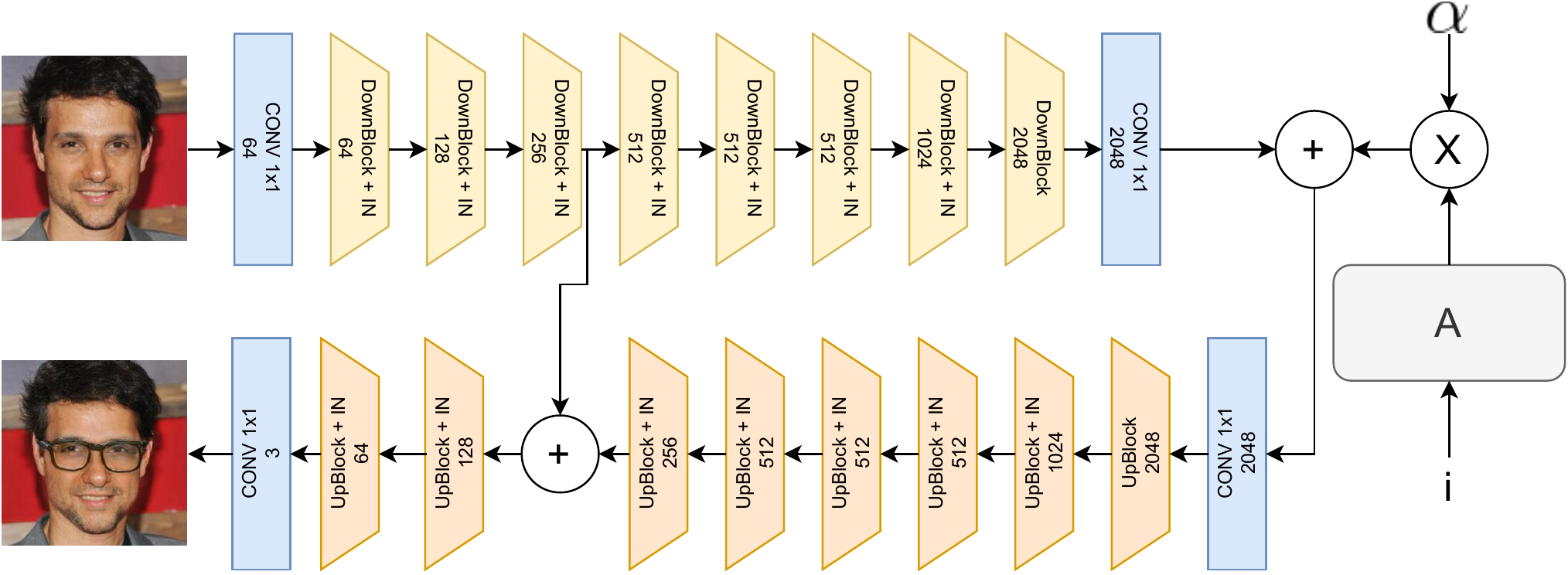}
    \caption{Generator architecture. Numbers correspond to the output channels of each block.}
    \label{fig:sup_generator}
\end{figure}

\paragraph{Generator.}
Our generator is composed of an encoder and decoder as shown in Fig. \ref{fig:generator}.
For encoder, we use 8 successive blocks that perform downsampling which reduce feature map dimensions to 1x1. In our decoder, we have an architecture symmetric to encoder, which is composed of 8 successive upsampling blocks. Except the last downsampling block and the first upsampling block, we use instance normalization denoted as (+IN). The channels increase as \{64, 64, 128, 256, 512, 512, 512, 1024, 2048\} (for output resolution 256x256) in the encoder and decrease in a symmetric way in the decoder.
In addition to these building blocks, we use a skip connection between the encoder and decoder as shown in Fig. \ref{fig:sup_generator}.

\paragraph{Residual Blocks. }
Each DownBlock and UpBlock has a residual block with $3x3$ convolutional filters followed by a downsampling and upsampling layer, respectively. For downsampling, we use average pooling and for upsampling, we use nearest-neighbor.
We use LeakyReLU activation layer and instance normalization layer in each convolutional module.

\begin{figure}
    \centering
    \includegraphics[width=1.0\linewidth]{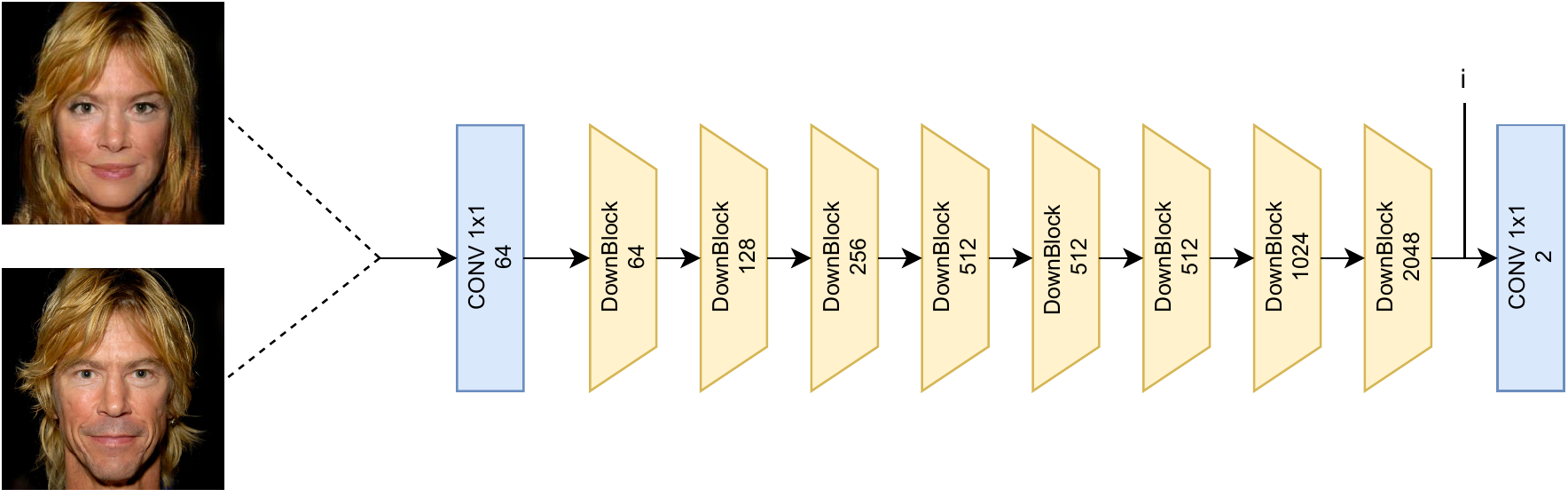}
    \caption{Architecture of the discriminator. Discriminator takes an input image and processes it with downsampling blocks with increased number of channels. Towards the end, the extracted feature map with 1x1 feature dimensions is concatenated with the scale of the input image. As we perform scale extraction for the image in the cycle-translation path, no additional scale extraction is needed.}
    \label{fig:dis}
\end{figure}

\paragraph{Discriminator.}  Discriminator also employs an architecture with decreasing resolution and increasing channel size as given in Fig. \ref{fig:dis}. Just like the generator, we build our discriminator with channel sizes of \{64, 64, 128, 256, 512, 512, 512, 1024, 2048\}, that reduces the feature map dimensions to 1x1. At the end, we concatenate the extracted style $\alpha_t$ from the input image to this latent code and apply a 1x1 convolution. This final convolution is specific to each tag-attribute pair so that the model can use this information.

\paragraph{Hyperparameters.}  For training our framework, we set the following parameters; $\lambda_{a} = 1$, $\lambda_{rec} = 1.5$, $\lambda_{s} = 1$, $\lambda_{o} = 1$ and $\lambda_{sp} = 0.05$. 
We use a  learning rate of $10^{-4}$ and train our model for 500K iterations with a batch size of 8 on a single GPU.
For the feature encoding and feature directions in matrix $A$, we use a 2048 dimensional vector representation same as the channel size of the last convolutional layer from the encoder.

\section{Additional Results}

\begin{figure}
    \centering
    \includegraphics[width=1.0\linewidth]{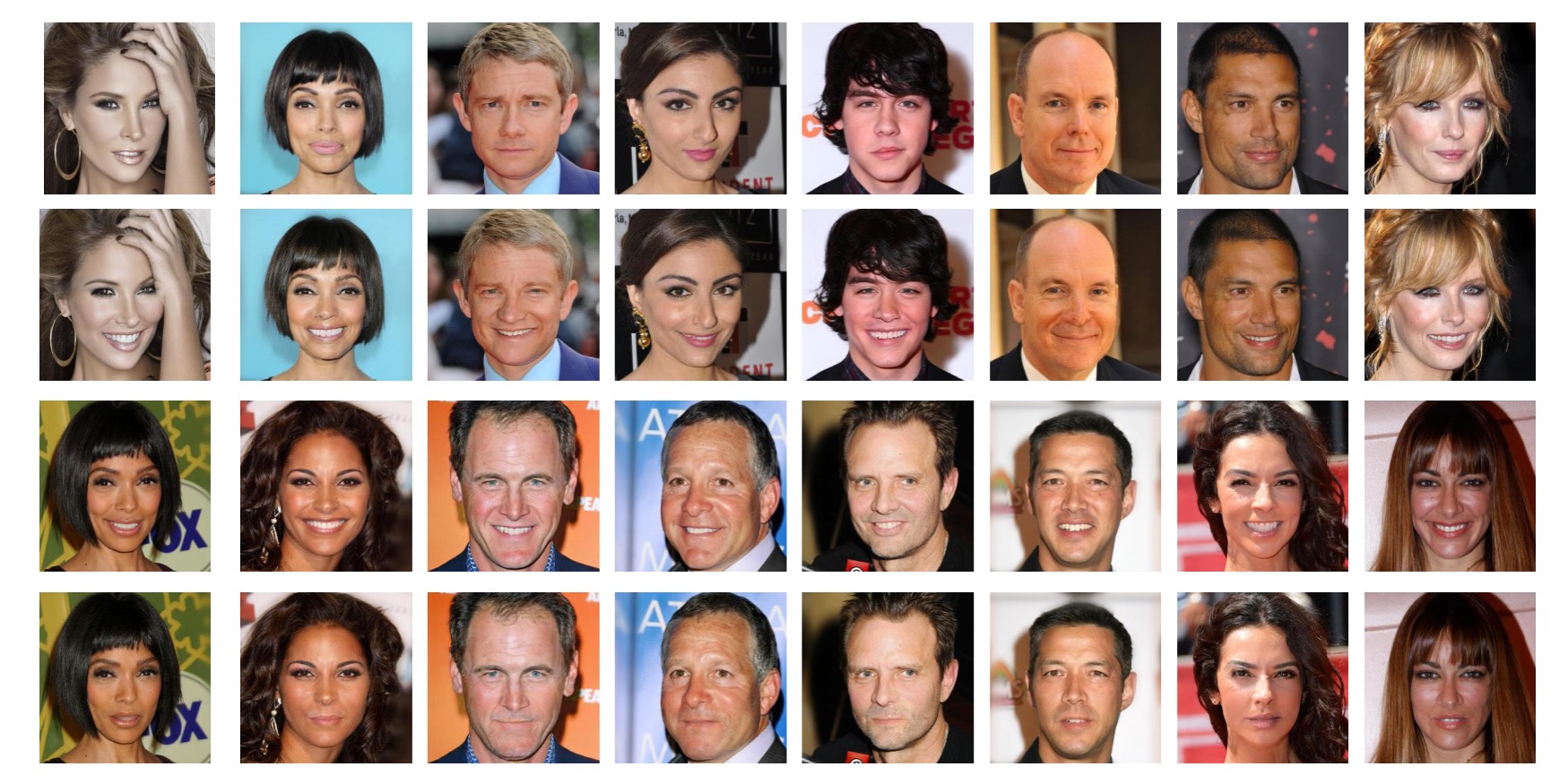}
    \caption{Smile tag manipulation results. First and third rows show input images. Second and forth rows show image translation results.}
    \label{fig:sup_smile}
\end{figure}

\begin{figure}
    \centering
    \includegraphics[width=1.0\linewidth]{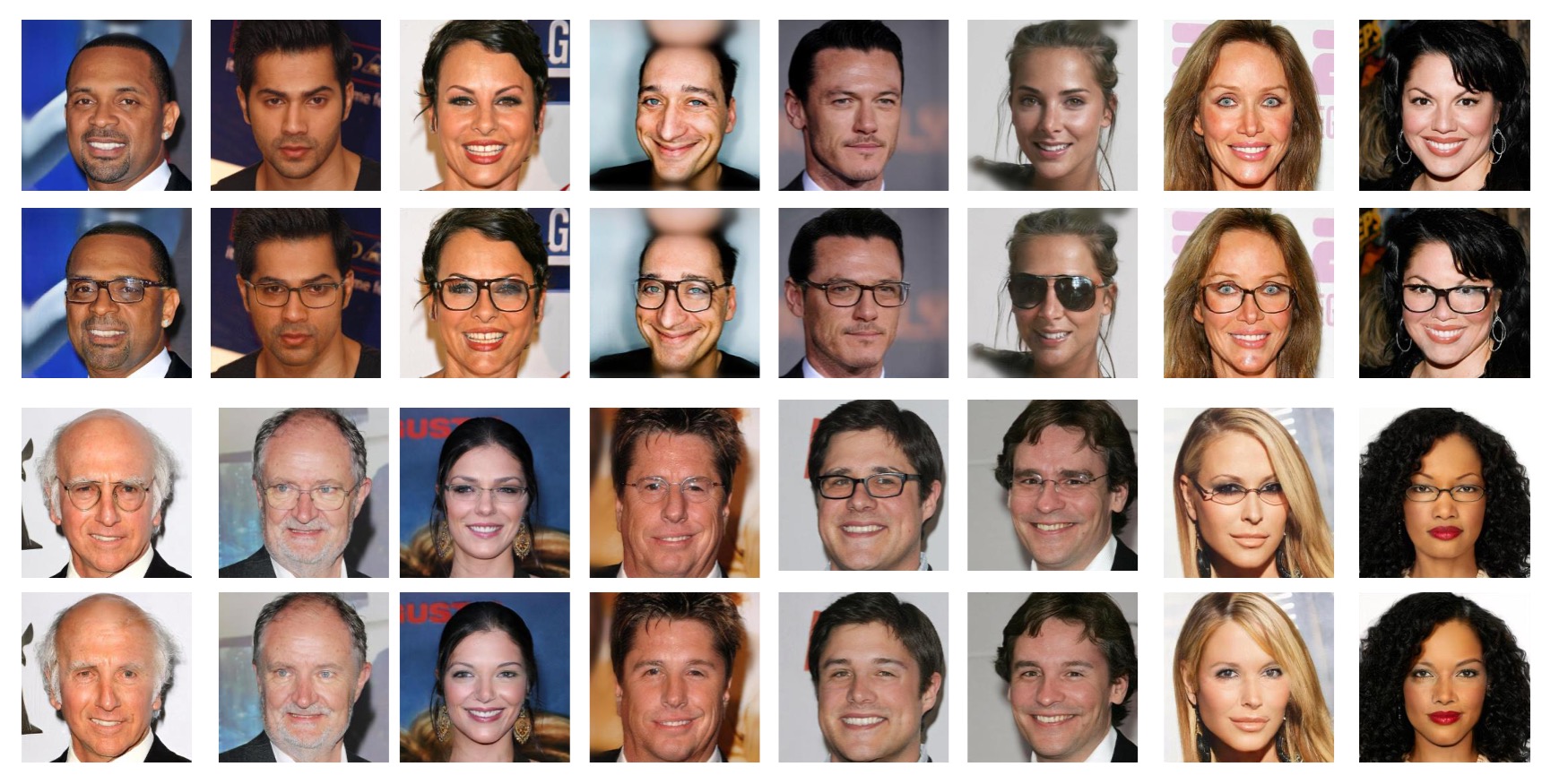}
    \caption{Glasses tag manipulation results. First and third rows show input images. Second and forth rows show image translation results.}
    \label{fig:sup_eyeglsses}
\end{figure}

\begin{figure}
    \centering
    \includegraphics[width=1.0\linewidth]{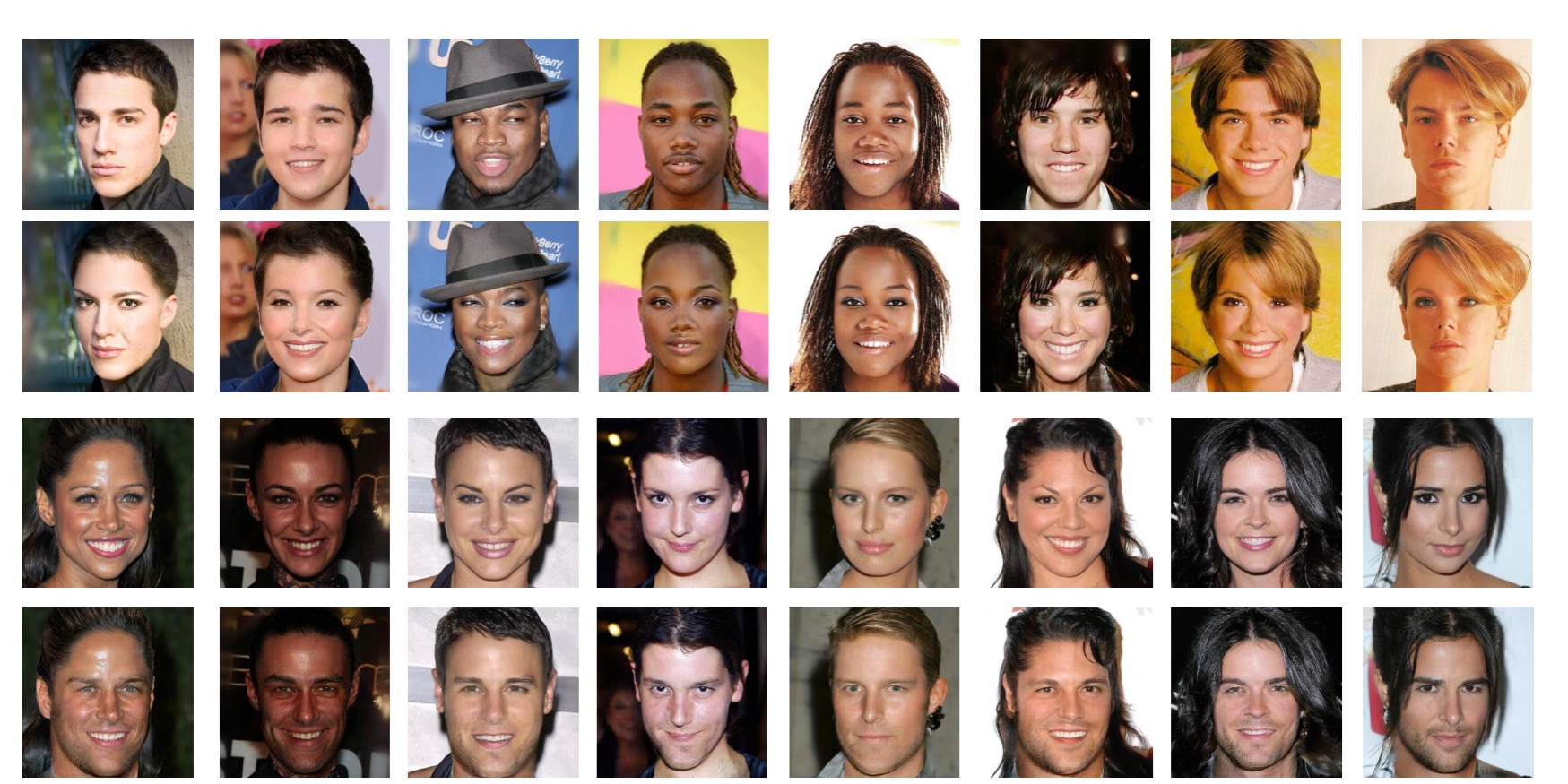}
    \caption{Gender tag  manipulation results. First and third rows show input images. Second and forth rows show image translation results.}
    \label{fig:sup_gender}
\end{figure}

\begin{figure}
    \centering
    \includegraphics[width=1.0\linewidth]{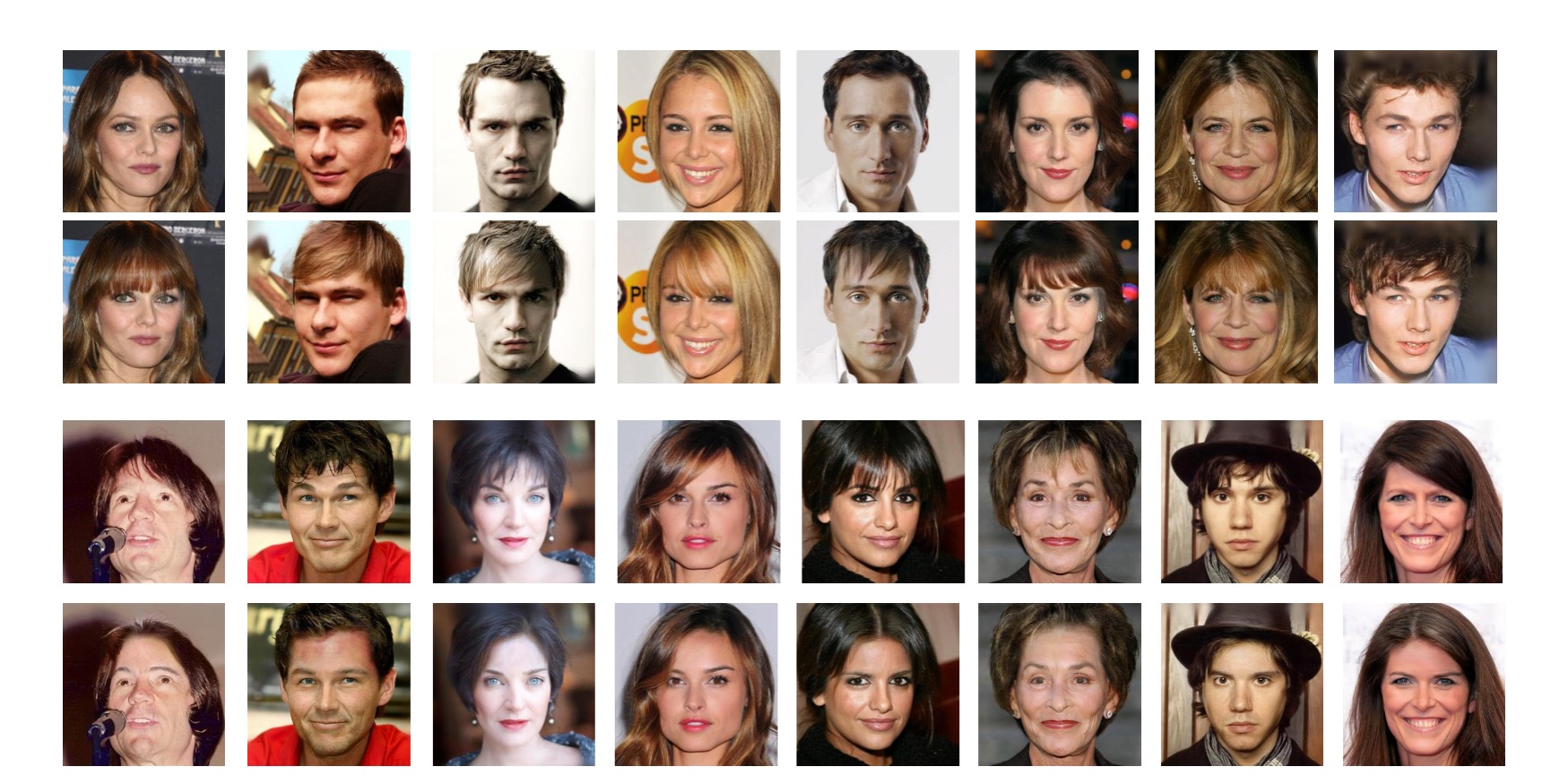}
    \caption{Bangs tag  manipulation results. First and third rows show input images. Second and forth rows show image translation results.}
    \label{fig:sup_bangs}
\end{figure}

\begin{figure}
    \centering
    \includegraphics[width=1.0\linewidth]{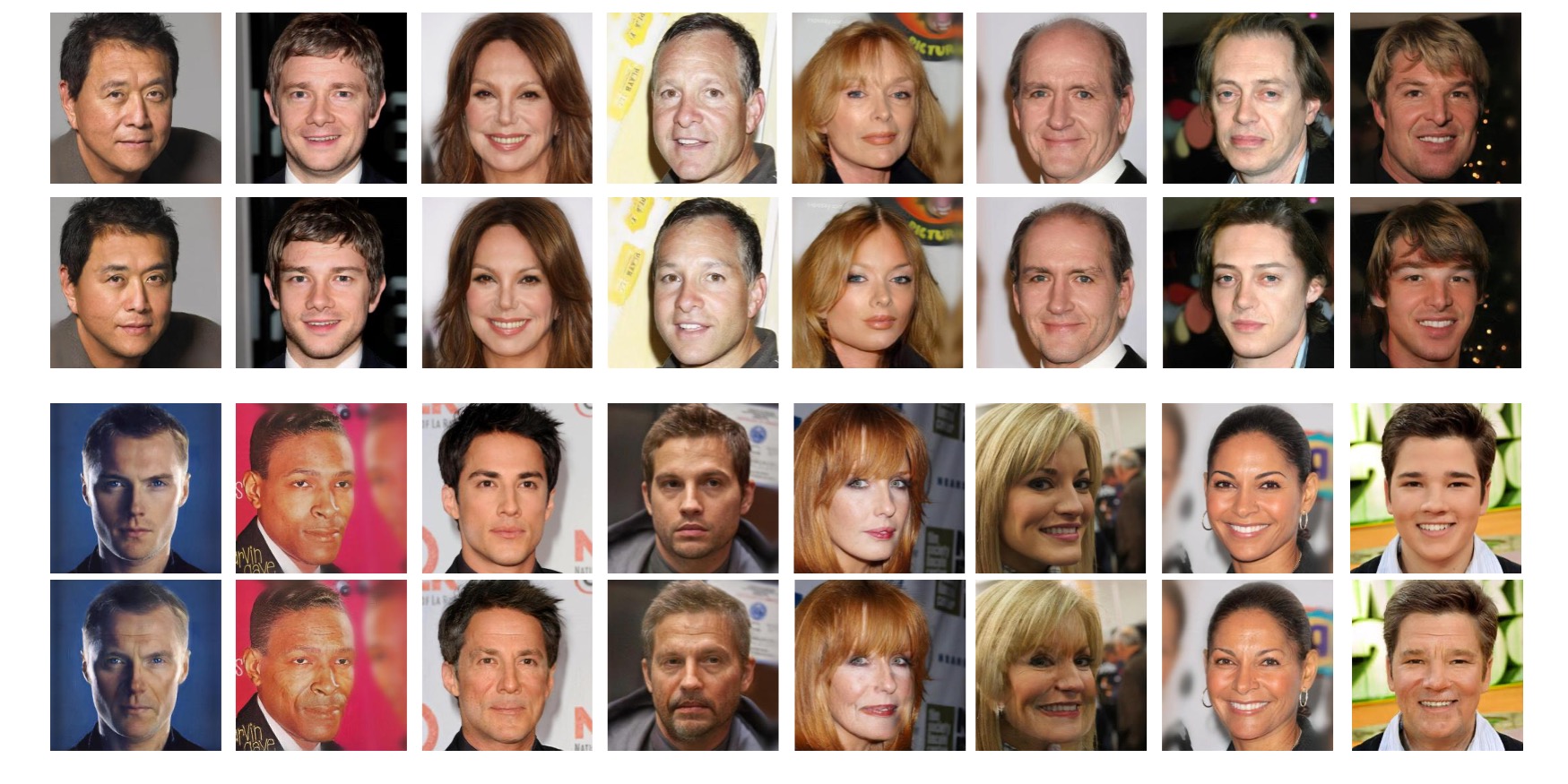}
    \caption{Age tag  manipulation results. First and third rows show input images. Second and forth rows show image translation results.}
    \label{fig:sup_age}
\end{figure}

\begin{figure}
    \centering
    \includegraphics[width=1.0\linewidth]{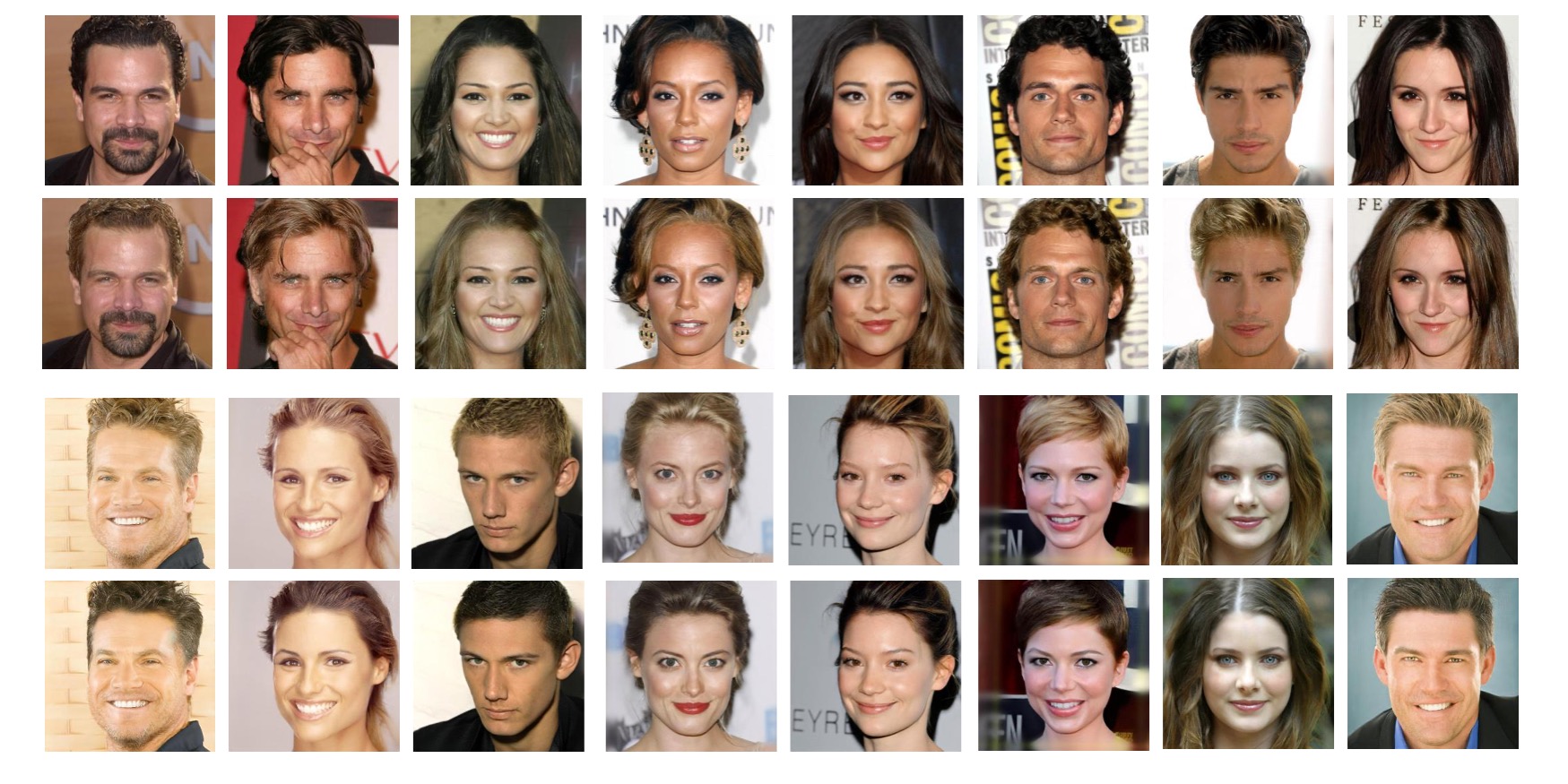}
    \caption{Hair tag  manipulation results. First and third rows show input images. Second and forth rows shows image translation results.}
    \label{fig:sup_hair}
\end{figure}

We provide additional qualitative results of our method in Fig. \ref{fig:sup_smile}, \ref{fig:sup_eyeglsses}, \ref{fig:sup_gender}, \ref{fig:sup_bangs}, \ref{fig:sup_age}, and \ref{fig:sup_hair}.

\newpage
\bibliographystyle{splncs04}
\bibliography{egbib}
\end{document}